# Precision Rehabilitation for Patients Post-Stroke based on Electronic Health Records and Machine Learning

**Fengyi Gao, MS[1], Xingyu Zhang, PhD[2], Sonish Sivarajkumar, BS[3], Parker E. Denny, DPT[4], Bayan M. Aldhahwani, PT, MS[4,5], Shyam Visweswaran, MD, PhD[3,6], Ryan Shi, PhD[7], William Hogan, MD, MS[8], Allyn Bove, DPT, PhD[4], Yanshan Wang, PhD[1,3,6,8,10]**

**[1]Department of Health Information Management, University of Pittsburgh, Pittsburgh, PA; [2]Department of Communication Science and Disorders, University of Pittsburgh, Pittsburgh, PA; [3]Intelligent Systems Program, School of Computing and Information, University of Pittsburgh, Pittsburgh, PA; [4]Department of Physical Therapy, University of Pittsburgh, Pittsburgh, PA; [5]Department of Medical Rehabilitation Sciences, Umm Al-Qura University, Makkah, Saudi Arabia; [6]Department of Biomedical Informatics, University of Pittsburgh, Pittsburgh, PA;[7]Department of Computer Science, University of Pittsburgh Medical Center, Pittsburgh, PA; [8]Data Science Institute, Medical College of Wisconsin, Milwaukee, WI; [9]Clinical and Translational Science Institute, University of Pittsburgh, Pittsburgh, PA; [10]Hillman Cancer Center, University of Pittsburgh Medical Center, Pittsburgh, PA;**

**Abstract**

**Objective** In this study, we utilized statistical analysis and machine learning methods to examine whether rehabilitation exercises can improve patients post-stroke functional abilities, as well as forecast the improvement in functional abilities. Our dataset is patients' rehabilitation exercises and demographic information recorded in the unstructured electronic health records (EHRs) data and free-text rehabilitation procedure notes. Through this study, our ultimate goal is to pinpoint the specific rehabilitation exercises that can effectively aid post-stroke patients in improving their functional outcomes in basic mobility (BM) and applied cognitive (AC) domains.

**Data sources** We collected data for 265 stroke patients from the University of Pittsburgh Medical Center, accessed through the Rehabilitation Datamart With Informatics iNfrastructure for Research (ReDWINE).

**Methods** We employed a pre-existing natural language processing (NLP) algorithm to extract data on rehabilitation exercises and developed a rule-based NLP algorithm to extract Activity Measure for Post-Acute Care (AM-PAC) scores, covering basic mobility (BM) and applied cognitive (AC) domains, from procedure notes. AM-PAC scores were collected at the initial rehabilitation visit and followed up at one and two months—key recovery periods. Changes in AM-PAC scores were classified based on the minimal clinically important difference (MCID), and significance was assessed using Friedman and Wilcoxon tests. To identify impactful exercises, we used Chi-square tests, Fisher's exact tests, and logistic regression for odds ratios. Additionally, we developed five machine learning models—logistic regression (LR), Adaboost (ADB), support vector machine (SVM), gradient boosting (GB), and random forest (RF)—to predict outcomes in functional ability.

**Results** Statistical analyses revealed significant associations between functional improvements and specific exercises. In the AC domain, the BALANCE exercise showed substantial early-stage improvement ($p<0.001$, OR=7.81, 95% CI 2.21-30.30). In the BM domain, significant late-stage improvements were noted with EYES CLOSED ($p<0.05$, OR=2.06, 95% CI 1.08-3.94) and GAIT exercises ($p<0.01$, OR=2.38, 95% CI 1.25-4.56). The Random Forest model achieved the best performance in predicting functional outcomes.

**Conclusion** In this study, we identified three rehabilitation exercises that significantly contributed to patient post-stroke functional ability improvement in the first two months. Additionally, the successful application of a machine learning model to predict patient-specific functional outcomes underscores the potential for precision rehabilitation.

## Introduction

The cerebrovascular accident (CVA) occurs when the blood supply to part of the brain is interrupted or reduced, resulting in brain damage and long-term disabilities such as paralysis and speech impairment[1]. As the leading cause of serious long-term disability in the United States, stroke affects more than 795,000 individuals each year, including 610,000 initial strokes and 185,000 recurrent strokes[2]. Recovery for patients post-stroke heavily relies on rehabilitation therapies to enhance functional abilities by targeting specific muscle groups, improving strength, range of motion and coordination, and ultimately improving their overall quality of life. However, a significant limitation of current rehabilitation programs is their generic nature. These therapies may not be fully effective for every individual since they do not account for the unique needs and variations in the condition of each patient. There is a growing need for personalized rehabilitation programs that are tailored specifically to the requirements of each individual to optimize function recovery outcomes.

Precision rehabilitation is a personalized approach that emphasizes individual-level functioning, aiming to deliver the right exercises, at the right time, to the right individual[3]. It enables the implementation of proactive and timely interventions that are specifically individualized for each person, leading to improved effectiveness and efficiency in delivering rehabilitation services[3]. In addition, in current rehabilitation therapies, one of the major challenges is the predominance of unstructured data fields, which contain a wealth of information regarding rehabilitation exercises. Without structured data, it is extremely difficult to analyze and extract meaningful insights. Furthermore, specific rehabilitation exercises are often embedded within these unstructured fields. Consequently, we cannot understand and evaluate the rationale for selecting particular rehabilitation exercises, resulting in a considerable obstacle to understanding and evaluating their effectiveness. Thus, utilizing advanced methods like Natural Language Processing (NLP) and Machine Learning (ML) is pivotal for precision rehabilitation. These cutting-edge methods allow physicians to analyze vast amounts of data efficiently, tailor treatment plans to individual needs, and predict patient post-stroke functional ability outcomes accurately. By leveraging NLP, physicians can extract valuable insights from unstructured electronic health records (EHRs) to inform decision-making processes[25]. Machine Learning methods enable the identification of patterns and trends within patient data, facilitating personalized rehabilitation strategies that optimize recovery outcomes[26]. Ultimately, these advanced methods enhance the precision and efficacy of rehabilitation exercises, empowering patients to achieve better functional ability outcomes and quality of life.

The widespread adoption of EHRs enables the access to the data for large populations that could facilitate precision rehabilitation analytics. However, a significant challenge exists: most crucial details about rehabilitation exercises are documented in unstructured EHRs, particularly within rehabilitation procedure notes, which complicates data analysis. Thus, in our previous study[9], we have developed NLP algorithms to extract detailed rehabilitation exercise information, which includes the type of motion, side of the body, location on the body, the plane of motion, duration, information on sets and reps, exercise purpose, exercise type, and body position. Building on the findings of the previous study, this study seeks to utilize the rehabilitation exercise information extracted from notes and other structured EHR data to 1) analyze the efficacy of various rehabilitation exercises on patient post-stroke functional ability outcomes, and 2) develop machine learning models to predict patient post-stroke functional ability outcomes given prescribed rehabilitation therapies.

**Related Work**

For post-stroke patients, rehabilitation exercises consist primarily of resistance exercises, aerobic exercises, and functional exercises. Strengthening muscles and improving motor function are the main objectives of resistance exercises. Studies have shown that resistance exercises that are isometric shoulder exercises with a sling and wrist strengthening programs enhance upper limb motor function significantly[11]. The purpose of aerobic exercises is to improve cardiovascular health and endurance. Lower extremity aerobic exercises, including treadmill training with or without added load, have shown improvements in gait speed and weight-bearing symmetry[12]. Upper extremity aerobic exercises, such as arm ergometers, have also proven effective for stroke patients[13]. Functional exercises involve activities that simulate daily tasks and movements. This includes robot-assisted therapy and electrical stimulation devices for home care[14,15]. The majority of prior studies focus on these three types of rehabilitation exercises separately, but our research covers a wide range of exercises, such as balance exercises, flexibility exercises and strength exercises, etc.

Rehabilitation exercises for patients post-stroke at different stages require a tailored approach to maximize outcomes and promote recovery. Studies have defined early subacute as a period of less than three months, late subacute as a period of three to six months, and chronic as a period of more than six months[16]. It has been shown that equipment-assisted treatment or resistance-combined exercises can help patients achieve better outcomes during the early subacute stage[17,18]. Patients at the Late Subacute Stage may benefit from home-based exercises, including resistance training and functional task practice[19,20]. In the chronic stage, patients can benefit from combining functional, aerobic, and resistance exercises in complex exercise programs[21]. Thus, stroke patients should be evaluated according to their current stage of recovery in a comprehensive rehabilitation program. We examined exercises that improve functional outcomes in stroke patients within two months of their stroke, a period that appears to be the most effective for stroke rehabilitation.

With the application of machine learning methods to stroke rehabilitation, it is possible to predict the functional outcomes of patients post-stroke and aid physicians in making decisions to maximize patients post-stroke functional outcomes. In many studies, random forest and SVM methods are used to predict patients post-stroke functional outcomes and achieve good performance[22,23,24]. We developed a total of five binary classification models based on these studies in order to predict the improvement of functional recovery of patients post-stroke, especially in the first two months.

## Methods

### *Data Collection*

We identified a cohort of 265 patients diagnosed with stroke at the University of Pittsburgh Medical Center (UPMC) and retrieved their EHR records across the outpatient rehabilitation settings from the Rehabilitation Datamart With Informatics iNfrastructure for rEsearch (ReDWINE) data warehouse[4]. The patients in our study had an average age of 64 years. The detailed demographics of this cohort can be found in the previous study[9].

### *Function Outcome Extraction*

Accurately assessing the functional ability outcome of patients post-stroke is crucial when evaluating the effectiveness of rehabilitation exercises. In the UPMC Rehabilitation Institute clinics, the Activity Measure for Post-Acute Care (AMPAC) is routinely collected at regular intervals for all patients receiving outpatient rehabilitation. The AMPAC quantitatively measures functional abilities and activity limitations in individuals undergoing post-acute care[5] in various domains, including basic mobility (BM) and applied cognitive (AC). More specifically, BM assesses physical movements and mobility skills, and AC evaluates cognitive functions required for practical tasks. By assessing these domains, healthcare professionals can comprehensively understand a patient's functional abilities and limitations, which can help guide treatment planning, rehabilitation, and care management. In our dataset, the AMPAC score is recorded in the free-text rehabilitation procedure notes with a clear pattern. Therefore, by using regular expressions, we extracted BM and AC scores from the patients' first visit, at 1 month (1M), and at 2 months (2M).

To determine changes at the individual level during both Early (from first visit to 1M) and Late (from 1M to 2M) Stages. Patients post-stroke have different rehabilitation interventions at different stages. The first three months after stroke are early subacute. This stage is a crucial time for rehabilitation intervention, as it can minimize disability and prevent complications. So we focus on Early and Late Stages to identify what specific rehabilitation exercises are most effective in improving functional recovery for patients post-stroke. In addition, patients were categorized into "improved" and "not improved" through the minimal clinically important difference (MCID). The MCID is defined as the minimal change in a score that is important to the patient[6]. Since no specific MCID for patients post-stroke has been reported for the AM-PAC, we estimated the MCID based on a gain or loss of 0.2 standard deviations (SD) from the pooled scores of the first visit, 1M, and 2M[7,8]. If the patient's AM-PAC score exceeds the estimated MCID, it is considered as improved; otherwise, it is classified as not improved.

*Statistical Analysis*

To assess whether the improvements in patients' functional abilities were significant over time, we utilized nonparametric methods suited for the non-normal distribution of AM-PAC scores. Specifically, we applied the Friedman test to evaluate differences across three time points—initial visit, one month, and two months—in both the basic mobility (BM) and applied cognitive (AC) domains. Additionally, the Wilcoxon signed-rank test was used to identify significant changes between specific intervals: from the initial visit to one month (Early Stage), from one month to two months (Late Stage), and from the initial visit to two months.

Sequential statistical tests were carried out to determine which rehabilitation exercises significantly improve functional abilities in post-stroke patients. Initially, we utilized Chi-square and Fisher's exact tests to explore the association between specific exercises and functional improvements. In the contingency table, each cell represents the number of patients corresponding to different categories (e.g., those who improved and participated in the BALANCE exercise). If a cell contained fewer than five patients, Fisher's exact test was employed; otherwise, the Chi-square test was used. We retained the calculated p-values from these tests. Considering the range of specific rehabilitation exercises, we set a p-value threshold of 0.3 to focus on the most relevant exercises for our study. Subsequently, we computed the odds ratios and 95% confidence intervals for these selected exercises, reflecting the likelihood of improvement among patients who underwent a particular exercise compared to those who did not.

*Machine Learning Models*

The input of the model contains a set of 115 specific rehabilitation exercises and three demographic features, including sex, age and race, which are extracted from EHRs. The model prediction target is the binary patient’s outcome: improved or not-improved.

In our study, we have developed a total of five binary classification models with the aim of predicting the improvement of patient post-stroke functional ability outcomes within the Early and Late Stages in each domain. Due to the limited dataset at our disposal and the infeasibility of employing deep learning techniques, our recourse is to rely on conventional machine learning approaches grounded in mathematical principles. The five models are logistic regression (LR), Adaboost (ADB), support vector machine (SVM), gradient boosting (GB), and random forest (RF). To evaluate the performance of each model, we utilize three metrics: F1 score, precision, recall, AUC and accuracy. These metrics provide valuable insights into the effectiveness of the models in predicting functional improvement.

We employed a 3-fold cross-validation method to enhance our model's performance assessment reliability over a single train-test split. To address class (improved vs. not improved) imbalance, where the improved class can be as low as 15%, we applied over-sampling to the training dataset to augment minority class instances by duplicating existing samples. This approach ensures that the model is equally trained on both minority and majority classes, thus promoting unbiased learning and improving overall performance. For evaluation, we adopted a weighted approach to the metrics (e.g. precision, recall, F1 score) , which emphasizes the ratio of improved to not-improved instances, where weight is the inversely proportional to class frequencies. This weighting method makes the evaluation more attuned to the model's capability of detecting positive changes or enhancements. This could be crucial depending on the specific application or problem domain. We constructed a ROC curve based on the cross-validation outcomes to visually represent and gauge the model's discriminatory power between classes.

## Results

### *Results* of *Statistical Analysis*

Through comparing the average AM-PAC scores across various periods, the Friedman test , a non-parametric statistical test, reveals statistical significance for all domains ($p < .05$) across the first visit, 1M and 2M. Furthermore, the Wilcoxon signed-rank test, another non-parametric test used for paired samples, indicates a significant improvement ($p < .05$) in all domains during the initial two months, as shown in Fig. 1.

**Figure 1.** Non-parametric statistical results for two domains at three different time points.

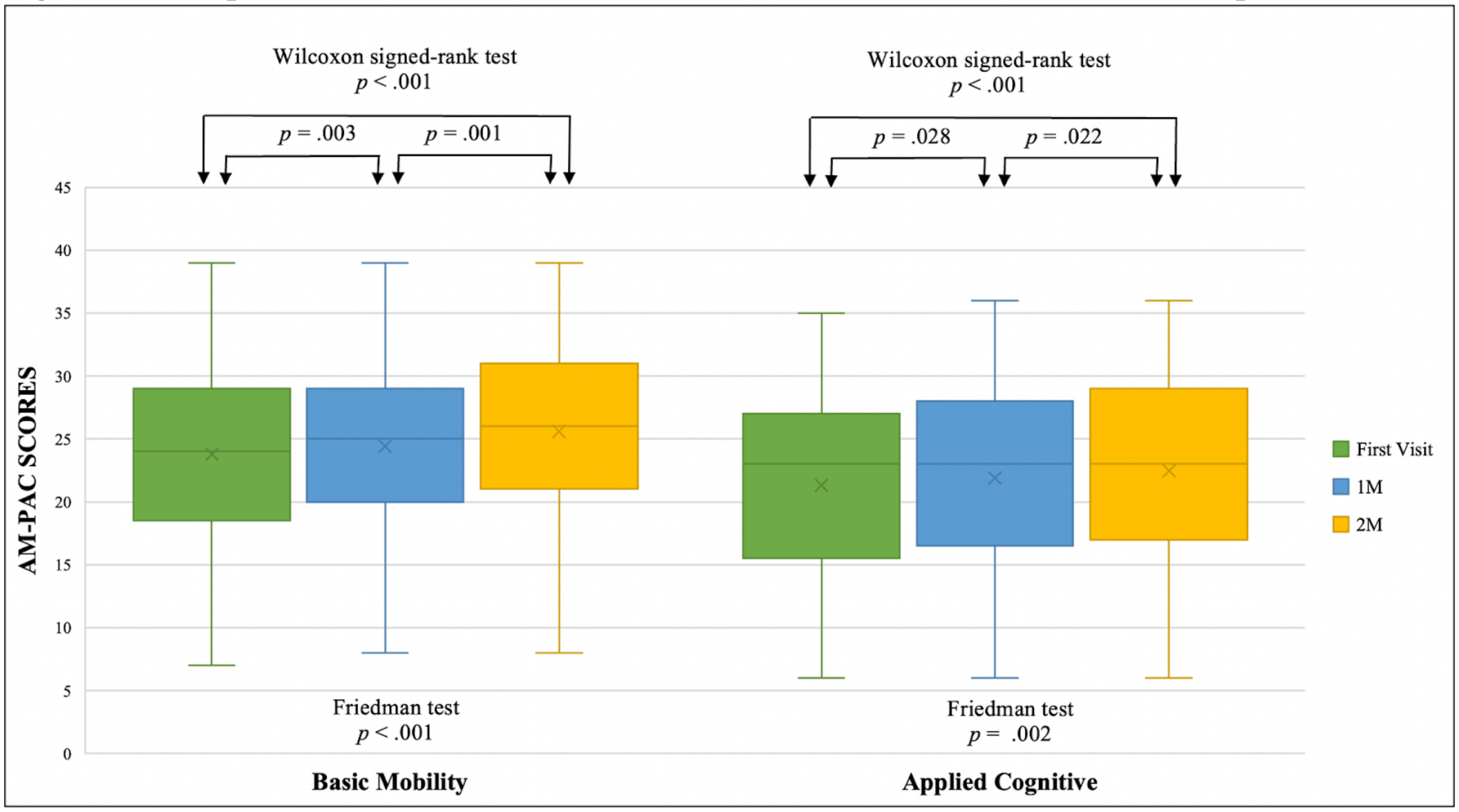

**Note:** The box plot shows the mean (X) and median (line) for two domains. The p value below the plots represents the Friedman test results used for each domain in order to examine significant differences among the three time points. The p values above the plots correspond to Wilcoxon signed-rank tests between two time points.

Among the 201 patients in the BM domain, 31 demonstrated important improvement at the Early Stage while 170 did not. However, no exercise demonstrated significant association with improvement as shown in Table 1. In the meanwhile, 66 patients demonstrated important improvement at the Late Stage while 135 did not. At the Late Stage, EYE CLOSED (p-value = 0.021, OR=2.06, 95% CI 1.08-3.94) and GAIT (p-value = 0.006, OR=2.38, 95% CI 1.25-4.56) presented significant contribution to the important improvement as shown in Table 2. This indicates that incorporating more EYE CLOSED or GAIT exercises could benefit patients' improvement in BM domain during the Late Stage, specifically from the 1M to 2M.

Similarly, out of the 109 patients, 19 showed significant improvement in the AC domain during the Early Stage, and 22 during the Late Stage, whereas 93 and 87 patients did not exhibit notable improvement at these respective stages. During the Early Stage, as shown in Table 3, there is significant association between BALANCE exercise and improvement of AC domain (p-value <0.001, OR=7.81, 95% CI 2.21-30.30). In the meanwhile, FOAM (p-value = 0.048, OR=3.29, 95% CI 0.98-12.11) and TANDEM (p-value =0.025, OR=3.98, 95% CI 1.00-14.99) presented marginal significance to such improvement. For the Late Stage, only BICEP CURLS (p-value =0.106, OR=2.38, 95% CI 1.25-4.56) presented marginal significance to the improvement of AC domain, but no exercise demonstrated significant contribution as shown in Table 4.

Overall, the statistical analysis indicated that a significantly higher number of patients showed improvement in the AC/BM domains during the first two months, and certain exercises played a significant role in facilitating this improvement.

**Table 1** Rehabilitation exercises and demographic features with improvement status in BM domain at Early Stage

| Category | Type | Improved N(%) 31 (15%) | Not-Improved N(%) 170 (85%) | *p* value | Odds Ratio |
|---|---|---|---|---|---|
| SEX | FEMALE | 17 (55%) | 79 (46%) | 0.508 | 1.40 (0.60-3.27) |
| | MALE | 14 (45%) | 91 (54%) | | |
| RACE | WHITE | 25 (81%) | 138 (81%) | 0.999 | 0.97 (0.35-3.12) |
| | NOT_WHITE | 6 (19%) | 32 (19%) | | |
| AGE | < 40 | 0 (0%) | 13 (8%) | 0.134 | 0.00 (0.00-1.77) |
| | 40 - 60 | 12 (39%) | 44 (26%) | | 1.80 (0.73-4.29) |
| | > 60 | 19 (61%) | 113 (66%) | | 0.80 (0.34-1.94) |
| SEX | FEMALE | 17 (55%) | 79 (46%) | 0.508 | 1.40 (0.60-3.27) |
| | MALE | 14 (45%) | 91 (54%) | | |
| FEET TOGETHER | YES | 8 (26%) | 22 (13%) | 0.095 | 2.33 (0.80-6.27) |
| | NO | 23 (74%) | 148 (87%) | | |
| LUNGES | YES | 4 (13%) | 7 (4%) | 0.070 | 3.42 (0.69-14.56) |
| | NO | 27 (87%) | 163 (96%) | | |
| LIFTING | YES | 3 (10%) | 5 (3%) | 0.108 | 3.50 (0.52-19.20) |
| | NO | 28 (90%) | 165 (97%) | | |
| TRANSFER | YES | 3 (10%) | 7 (4%) | 0.187 | 2.48 (0.39-11.69) |
| | NO | 28 (90%) | 163 (96%) | | |
| FEET APART | YES | 7 (23%) | 25 (15%) | 0.288 | 1.69 (0.55-4.61) |
| | NO | 24 (77%) | 145 (85%) | | |
| MINI SQUATS | YES | 7 (23%) | 23 (14%) | 0.269 | 1.86 (0.61-5.13) |
| | NO | 24 (77%) | 147 (86%) | | |
| ROWS | YES | 1 (03%) | 21 (12%) | 0.210 | 0.24 (0.01-1.60) |
| | NO | 30 (97%) | 149 (88%) | | |
| SLIDING | YES | 4 (13%) | 41 (24%) | 0.241 | 0.47 (0.11-1.46) |
| | NO | 27 (87%) | 129 (76%) | | |

**Note:** We retained only the most relevant specific rehabilitation exercises by setting a p-value threshold of greater than 0.3.

**Table 2** Rehabilitation exercises and demographic features with improvement status in BM domain at Late Stage

| Category | Type | Improved N(%) 66 (33%) | Not-Improved N(%) 135 (67%) | *p* value | Odds Ratio |
|---|---|---|---|---|---|
| SEX | FEMALE | 26 (39%) | 70 (52%) | 0.131 | 0.61 (0.32-1.14) |
| | MALE | 40 (61%) | 65 (48%) | | |
| RACE | WHITE | 55 (83%) | 108 (80%) | 0.708 | 1.25 (0.55-3.01) |
| | NOT_WHITE | 11 (17%) | 27 (20%) | | |
| AGE | < 40 | 3 (5%) | 10 (7%) | 0.556 | 0.60 (0.10-2.43) |
| | 40 - 60 | 21 (32%) | 35 (26%) | | 1.33 (0.66-2.66) |
| | > 60 | 42 (64%) | 90 (67%) | | 0.88 (0.45-1.71) |
| BIOSTEP | YES | 3 (5%) | 1 (1%) | 0.202 | 6.32 (0.50-337.41) |
| | NO | 63 (95%) | 134 (99%) | | |
| **EYES CLOSED** | **YES** | **33 (50%)** | **44 (33%)** | **0.021** | **2.06 (1.08-3.94)** |
| | **NO** | **33 (50%)** | **91 (67%)** | | |
| FEET TOGETHER | YES | 16 (24%) | 18 (13%) | 0.071 | 2.07 (0.91-4.70) |
| | NO | 50 (76%) | 117 (87%) | | |
| STEP DOWN | YES | 7 (11%) | 6 (4%) | 0.126 | 2.54 (0.70-9.57) |
| | NO | 59 (89%) | 129 (96%) | | |
| WEIGHT BEARING | YES | 7 (11%) | 8 (6%) | 0.260 | 1.88 (0.55-6.24) |
| | NO | 59 (89%) | 127 (94%) | | |
| ARM BIKE | YES | 4 (6%) | 1 (1%) | 0.073 | 8.55 (0.82-428.83) |
| | NO | 62 (94%) | 134 (99%) | | |
| STEP UP | YES | 42 (64%) | 72 (53%) | 0.176 | 1.53 (0.80-2.95) |
| | NO | 24 (36%) | 63 (47%) | | |
| TAPPING | YES | 16 (24%) | 19 (14%) | 0.079 | 1.95 (0.86-4.37) |
| | NO | 50 (76%) | 116 (86%) | | |
| TOE RAISES | YES | 2 (3%) | 11 (8%) | 0.228 | 0.35 (0.04-1.69) |
| | NO | 64 (97%) | 124 (92%) | | |
| TREADMILL | YES | 11 (17%) | 15 (11%) | 0.272 | 1.60 (0.62-4.00) |
| | NO | 55 (83%) | 120 (89%) | | |
| CHEST PRESS | YES | 3 (5%) | 1 (1%) | 0.202 | 6.32 (0.50-337.41) |
| | NO | 63 (95%) | 134 (99%) | | |
| STEPPING | YES | 24 (36%) | 37 (27%) | 0.197 | 1.51 (0.76-2.96) |
| | NO | 42 (64%) | 98 (73%) | | |
| SIDE STEPPING | YES | 36 (55%) | 59 (44%) | 0.176 | 1.54 (0.82-2.92) |
| | NO | 30 (45%) | 76 (56%) | | |
| CARRYING | YES | 7 (11%) | 6 (4%) | 0.126 | 2.54 (0.70-9.57) |
| | NO | 59 (89%) | 129 (96%) | | |
| LONG ARC QUAD | YES | 19 (29%) | 24 (18%) | 0.099 | 1.86 (0.88-3.94) |
| | NO | 47 (71%) | 111 (82%) | | |
| LUNGE | YES | 11 (17%) | 14 (10%) | 0.255 | 1.72 (0.66-4.39) |
| | NO | 55 (83%) | 121 (90%) | | |
| FOAM | YES | 41 (62%) | 67 (50%) | 0.100 | 1.66 (0.88-3.19) |
| | NO | 25 (38%) | 68 (50%) | | |
| KNEE EXTENSION | YES | 7 (11%) | 8 (06%) | 0.260 | 1.88 (0.55-6.24) |
| | NO | 59 (89%) | 127 (94%) | | |
| FEET APART | YES | 17 (26%) | 20 (15%) | 0.080 | 1.99 (0.90-4.39) |
| | NO | 49 (74%) | 115 (85%) | | |
| PASSIVE ROM | YES | 9 (14%) | 10 (07%) | 0.199 | 1.97 (0.67-5.72) |
| | NO | 57 (86%) | 125 (93%) | | |
| **GAIT** | **YES** | **37 (56%)** | **47 (35%)** | **0.006** | **2.38 (1.25-4.56)** |
| | **NO** | **29 (44%)** | **88 (65%)** | | |

**Note:** We retained only the most relevant specific rehabilitation exercises by setting a p-value threshold of greater than 0.3.

**Table 3** Rehabilitation exercises and demographic features with improvement status in AC domain at Early Stage

| Category | Type | Improved N(%) 16 (15%) | Not-Improved N(%) 93 (85%) | *p* value | Odds Ratio |
|---|---|---|---|---|---|
| SEX | FEMALE | 7 (44%) | 46 (49%) | 0.880 | 0.80 (0.23-2.63) |
| | MALE | 9 (56%) | 47 (51%) | | |
| RACE | WHITE | 13 (81%) | 74 (80%) | 0.999 | 1.11 (0.27-6.69) |
| | NOT_WHITE | 3 (19%) | 19 (20%) | | |
| AGE | < 40 | 2 (12%) | 4 (4%) | 0.173 | 3.13 (0.26-24.34) |
| | 40 - 60 | 3 (19%) | 36 (39%) | | 0.37 (0.06-1.47) |
| | > 60 | 11 (69%) | 53 (57%) | | 1.65 (0.48-6.57) |
| GLIDE | YES | 2 (12%) | 5 (5%) | 0.273 | 2.49 (0.22-17.13) |
| | NO | 14 (88%) | 88 (95%) | | |
| EYES CLOSED | YES | 7 (44%) | 21 (23%) | 0.117 | 2.64 (0.74-9.11) |
| | NO | 9 (56%) | 72 (77%) | | |
| FOAM | YES | 10 (62%) | 31 (33%) | 0.048 | 3.29 (0.98-12.11) |
| | NO | 6 (38%) | 62 (67%) | | |
| SIT TO STAND | YES | 9 (56%) | 37 (40%) | 0.276 | 1.93 (0.58-6.70) |
| | NO | 7 (44%) | 56 (60%) | | |
| **BALANCE** | **YES** | **10 (62%)** | **16 (17%)** | **< 0.001** | **7.81 (2.21-30.30)** |
| | **NO** | **6 (38%)** | **77 (83%)** | | |
| STEP UP | YES | 9 (56%) | 37 (40%) | 0.276 | 1.93 (0.58-6.70) |
| | NO | 7 (44%) | 56 (60%) | | |
| STEP OVER | YES | 7 (44%) | 21 (23%) | 0.117 | 2.64 (0.74-9.11) |
| | NO | 9 (56%) | 72 (77%) | | |
| TANDEM | YES | 6 (38%) | 12 (13%) | 0.025 | 3.98 (1.00-14.99) |
| | NO | 10 (62%) | 81 (87%) | | |
| AMBULATION | YES | 6 (38%) | 20 (22%) | 0.205 | 2.17 (0.58-7.59) |
| | NO | 10 (62%) | 73 (78%) | | |
| PLANKS | YES | 2 (12%) | 2 (2%) | 0.189 | 6.32 (0.43-93.80) |
| | NO | 14 (88%) | 91 (98%) | | |
| EYES OPEN | YES | 5 (31%) | 14 (15%) | 0.150 | 2.54 (0.60-9.56) |
| | NO | 11 (69%) | 79 (85%) | | |

**Note:** We retained only the most relevant specific rehabilitation exercises by setting a p-value threshold of greater than 0.3.

**Table 4** Rehabilitation exercises and demographic features with improvement status in AC domain at Late Stage

| Category | Type | Improved N(%) 22 (20%) | Not-Improved N(%) 87 (80%) | *p* value | Odds Ratio |
|---|---|---|---|---|---|
| SEX | FEMALE | 11 (50%) | 42 (48%) | 0.999 | 0.61 (0.32-1.14) |
| | MALE | 11 (50%) | 45 (52%) | | |
| RACE | WHITE | 16 (73%) | 71 (82%) | 0.529 | 1.25 (0.55-3.01) |
| | NOT_WHITE | 6 (27%) | 16 (18%) | | |
| AGE | < 40 | 0 (0%) | 6 (7%) | 0.180 | 0.60 (0.10-2.43) |
| | 40 - 60 | 11 (50%) | 28 (32%) | | 1.33 (0.66-2.66) |
| | > 60 | 11 (50%) | 53 (61%) | | 0.88 (0.45-1.71) |
| TRUNK ROTATION | Yes | 3 (14%) | 3 (3%) | 0.177 | 6.32 (0.50-337.41) |
| | No | 19 (86%) | 84 (97%) | | |
| TREADMILL | Yes | 4 (18%) | 6 (7%) | 0.114 | 1.99 (0.90-4.39) |
| | No | 18 (82%) | 81 (93%) | | |
| TRICEPS | Yes | 3 (14%) | 4 (5%) | 0.290 | 8.55 (0.82-428.83) |
| | No | 19 (86%) | 83 (95%) | | |
| SIT TO STAND | Yes | 6 (27%) | 42 (48%) | 0.095 | 1.88 (0.55-6.24) |
| | No | 16 (73%) | 45 (52%) | | |
| FOAM | Yes | 11 (50%) | 29 (33%) | 0.215 | 1.95 (0.86-4.37) |
| | No | 11 (50%) | 58 (67%) | | |
| AMBULATION | Yes | 3 (14%) | 29 (33%) | 0.114 | 6.32 (0.50-337.41) |
| | No | 19 (86%) | 58 (67%) | | |
| BICEP CURLS | Yes | 6 (27%) | 11 (13%) | 0.106 | 2.38 (1.25-4.56) |
| | No | 16 (73%) | 76 (87%) | | |
| STEP OVER | Yes | 4 (18%) | 32 (37%) | 0.129 | 1.60 (0.62-4.00) |
| | No | 18 (82%) | 55 (63%) | | |

**Note:** We retained only the most relevant specific rehabilitation exercises by setting a p-value threshold of greater than 0.3.

*Results* of *Machine Learning*

Table 1 presents the model performance (F1 score, precision, recall, AUC and accuracy) for the two domains in both Early and Late Stages. In this study, we prioritized using the F1 score to select the best performing model. Among the models evaluated, RF demonstrated the highest performance for BM in both Early and Late Stages, as well as AC in both Early and Late Stages. These findings highlight the strengths of RF across different stages of evaluation metrics. In general, RF is the best performing model of the 5 models we developed, combining Early and Late Stages for each domain.

Based on the observation that RF is the best performing model, we presented its ROC curves, as shown in Fig. 2. The ROC curve is a graphical representation that illustrates the trade-off between the True Positive Rate (Sensitivity) and the False Positive Rate (1-Specificity) across various threshold values. It is a valuable tool for evaluating the performance of binary classification models. Furthermore, the AUC represents the overall performance of the model. Overall, the RF model has better performance on AC scores than BM scores, but none of them with AUC larger than 0.8, which is likely due to the imbalanced dataset.

**Table 1.** F1 score, precision, recall, AUC and accuracy of each machine learning model in the Early and Late Stages across each domain

| Domain | Machine Learning Models | Early Stage (0-1M) | | | | | Late Stage (1-2M) | | | | |
|---|---|---|---|---|---|---|---|---|---|---|---|
| | | Precision | Recall | F1 score | AUC | Accuracy | Precision | Recall | F1 score | AUC | Accuracy |
| BM | LR | 0.56 | 0.25 | 0.34 | 0.54 | 0.55 | 0.54 | 0.40 | 0.46 | 0.53 | 0.54 |
| | **RF** | **0.49** | **0.81** | **0.61** | **0.52** | **0.52** | **0.63** | **0.49** | **0.55** | **0.60** | **0.60** |
| | GB | 0.51 | 0.27 | 0.32 | 0.53 | 0.52 | 0.57 | 0.42 | 0.48 | 0.55 | 0.56 |
| | ADB | 0.58 | 0.37 | 0.43 | 0.56 | 0.56 | 0.58 | 0.48 | 0.52 | 0.57 | 0.57 |
| | SVM | 0.36 | 0.12 | 0.18 | 0.47 | 0.50 | 0.53 | 0.41 | 0.46 | 0.52 | 0.52 |
| AC | LR | 0.56 | 0.27 | 0.34 | 0.54 | 0.53 | 0.51 | 0.31 | 0.38 | 0.52 | 0.52 |
| | **RF** | **0.70** | **0.64** | **0.66** | **0.70** | **0.70** | **0.69** | **0.61** | **0.62** | **0.66** | **0.65** |
| | GB | 0.73 | 0.32 | 0.44 | 0.61 | 0.61 | 0.61 | 0.48 | 0.51 | 0.61 | 0.58 |
| | ADB | 0.67 | 0.42 | 0.52 | 0.62 | 0.63 | 0.56 | 0.72 | 0.63 | 0.61 | 0.61 |
| | SVM | 0.37 | 0.22 | 0.27 | 0.51 | 0.50 | 0.50 | 0.28 | 0.35 | 0.51 | 0.51 |

**Note**: LR: logistic regression, ADB: Adaboost, SVM: support vector machine, GB: gradient boosting, RF: random forest, BM: Basic Mobility, AC: Applied Cognitive

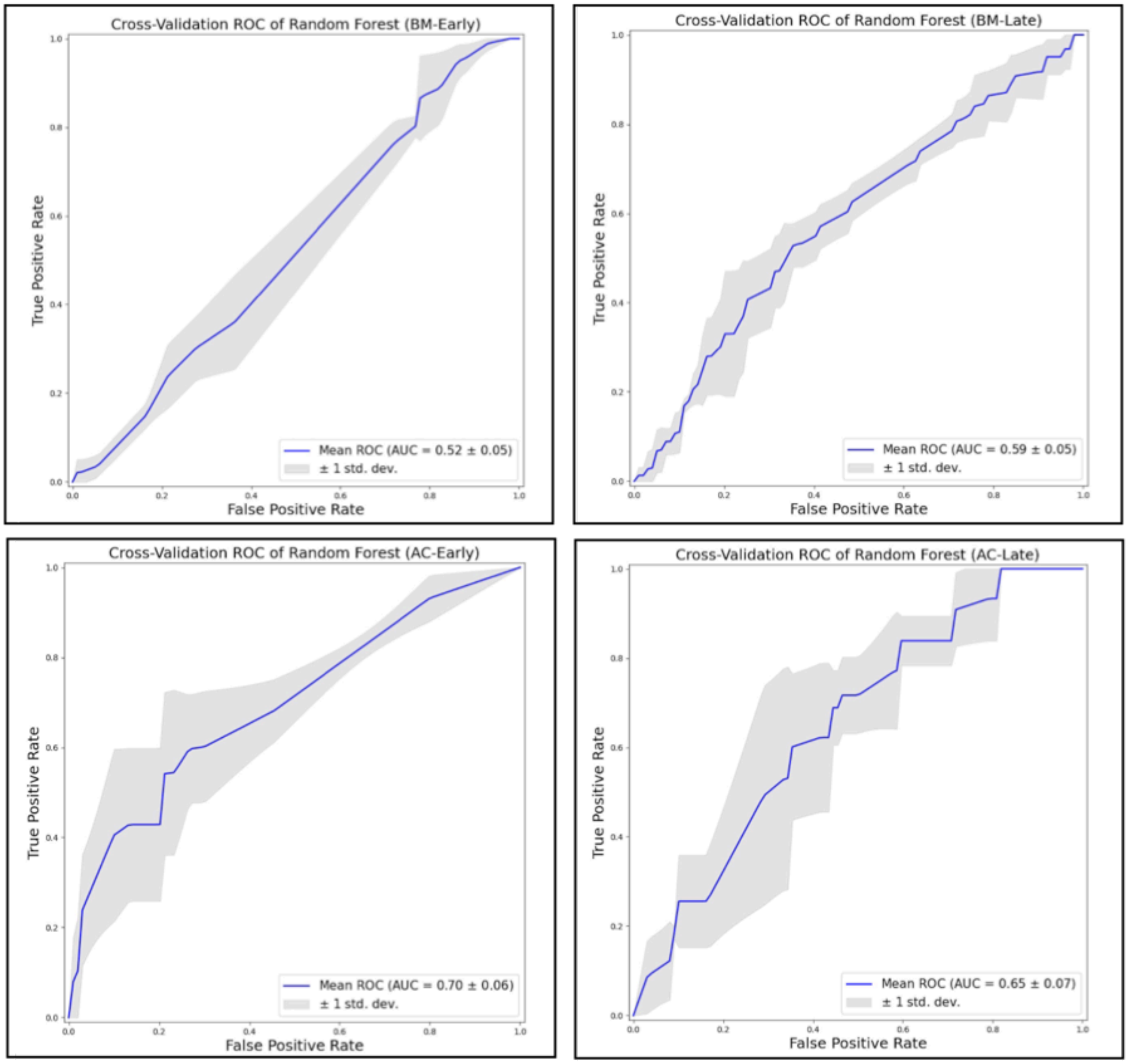

**Fig. 2** ROC curves for the random forest models across two domains and two stages (The blue line is the mean of three cross-validation ROC curve value and the gray shade is the standard deviation of three cross-validation ROC curve value)

## Discussion

*Important Rehabilitation Exercises*

Rehabilitation exercises facilitate patients' functional ability outcomes in the BM and AC domains for patients post-stroke, as demonstrated in our study. In particular, patients showed significant improvement during the early stages of rehabilitation, with certain rehabilitation exercises showing significant associations with these improvements.

Notably, EYE CLOSED and GAIT rehabilitation exercises could significantly benefit patients post-stroke functional ability outcomes in the BM domain at Late Stage. EYE CLOSED exercises improve balance, sensory integration, coordination, and motor control. Promotes neural plasticity, enhances balance, and coordinates. GAIT exercises can restore mobility, muscle strength, endurance, balance, and coordination for walking. Improves walking ability, enhances independence in daily living activities, and boosts quality of life for patients post-stroke.

Similarly, in the AC domain, our study identified significant associations between certain rehabilitation exercises and improvements in post-stroke functional ability outcomes during both the Early and Late Stages. During the Early Stage, BALANCE exercises exhibited a particularly strong association with improvement of patients post-stroke functional ability outcomes in the AC domain. This underscores that BALANCE exercises play a crucial role in rehabilitation therapy, especially during the Early Stages of rehabilitation. Focusing on BALANCE exercises can improve an individual's ability to maintain equilibrium and reduce falling risk.

However, it is noteworthy that fewer rehabilitation exercises demonstrated significant contributions to improvements in the AC domain at Late Stage and in the BM domain at Early Stage. This highlights the need for further research to identify additional interventions that can effectively patients post-stroke functional ability outcomes at these stages.

Overall, our findings emphasize the critical role of rehabilitation exercises in promoting patients post-stroke functional ability outcomes. Tailoring exercise interventions to enhance post-stroke functional ability outcomes during different stages could maximize therapeutic benefits and optimize patient outcomes. Furthermore, future research should investigate the mechanisms underlying combination exercises and identify novel interventions to improve patients post-stroke functional ability outcomes.

*Precision Rehabilitation*

By analyzing the data, we found that not all patients' functional ability had improved, which means that there are differences in functional ability improvement at the individual level in patients post-stroke undergoing rehabilitation. This indicates that a one-size-fits-all approach to stroke rehabilitation may not be optimal. Therefore, this requires our model to do personalized prediction to estimate whether or not the functional ability of an individual level will be improved in the future, so as to give physicians clinical decision support. Physicians can gain some insights from the model to create personalized exercises that address each patient's unique requirements and characteristics. By tailoring the precision rehabilitation interventions to suit the specific needs of each individual, healthcare professionals can maximize the potential for recovery and restoration of function.

## Acknowledgements

Research reported in this publication was supported by the School of Health and Rehabilitation Sciences Dean's Research and Development Award.

**Declaration of conflicting interests**

The author(s) declare no potential conflicts of interest with respect to the research, authorship, and/or publication of this article.

## Appendix

Rehabilitation Exercises from Rehabilitation Procedure Notes

| Categories of rehabilitation exercises | Key Phrases from Rehabilitation Procedure Notes |
|---|---|
| Upper Extremity Strength | Bicep curls, bicep(s)<br>Triceps, Tricep(s) extension, tricep(s) ext,<br>Chest press, bench press<br>Punch, punches<br>Overhead press<br>Shrug(s)<br>PNF |
| Lower Extremity Strength | Long arc quad, LAQ<br>Knee extension, Leg extension, leg ext<br>Step up<br>Step down<br>Toe raises), HR/TR<br>Minisquats<br>Lunge(s)<br>Weight bearing<br>Short arc quad, SAQ<br>Bridge(s), bridging<br>Glute set(s)<br>Clam(s), clamshell(s)<br>Hamstring curl, ham curl, HS curl<br>Heel raise(s), heel/toe raise(s)<br>Leg press, LP<br>Quad set(s), QS, quadset<br>Single leg raise, SLR |
| Trunk/Core Strength | Planks<br>Posterior pelvic tilt, PPT, pelvic tilt(s)<br>Pallof press, palof press<br>Transverse abdominus, TA, TrA, abdominal brace<br>Trunk extension<br>Curl up(s), sit up(s)<br>Chop(s)<br>Levator ani, LA<br>Deadbug(s)<br>Chin tucks, DNF, deep neck flex, chin retraction<br>Bird dog, Birddog<br>Cervical isometrics |
| Scapular Strength | Rows<br>Scapular retraction, scap retraction, scap squeezes<br>Serratus, SA<br>Ball on wall<br>Wing arm<br>Mid trap, MT<br>Low trap, LT<br>Pull down, lat pull down |
| Range of Motion | Sliding<br>Glide<br>Pulley(s)<br>Finger ladder<br>PROM<br>AAROM<br>Wand |
| Flexibility/Mobility | Trunk rotation, LTR, lower trunk ROT<br>Stretch, stretched, str |
| Balance/Vestibular | Balance<br>Tandem<br>Foam<br>Eyes open (EO)<br>Eyes closed (EC)<br>Feet together<br>Feet apart<br>Tap(s)<br>Weightshift, weight shift, weightshifting<br>Stance, SLS, semitandem<br>VORx1 |
| | Head turns, HT, HHT, VHT |
| Gait Training | Gait<br>Ambulation<br>Stepping<br>Sidestepping<br>Walk, walking<br>Stair(s) |
| Cardiovascular / Aerobic Training | Treadmill<br>Arm bike<br>Biostep<br>Bike, cycle<br>Nustep, Nu-step, Nu step<br>Endurance<br>METS<br>Elliptical, ET |
| Functional Mobility | Transfers<br>Sit to stand<br>Carrying<br>Step over<br>Lifting<br>Bed to chair, bed>chair, bed-chair<br>Bed mobility<br>Supine to sit, supine>sit, sup>sit<br>Rolling<br>Scoot, scooting<br>Door<br>Floor to stand, floor>stand, fl to stand, floor xfer, floor transfer |